\documentclass{article}

\usepackage[final,nonatbib]{neurips_2021}

\usepackage[utf8]{inputenc} 
\usepackage[T1]{fontenc}    
\usepackage{hyperref}
\hypersetup{colorlinks,allcolors=black}
\usepackage{url}            
\usepackage{booktabs}       
\usepackage{amsfonts}       
\usepackage{nicefrac}       
\usepackage{microtype}      
\usepackage{xcolor}         
\usepackage{graphicx}
\usepackage{xspace}
\usepackage{amssymb}
\makeatletter
\DeclareRobustCommand\onedot{\futurelet\@let@token\@onedot}
\def\@onedot{\ifx\@let@token.\else.\null\fi\xspace}
\def\eg{\emph{e.g}\onedot} 
\def\ie{\emph{i.e}\onedot}

\makeatother

\title{YMIR: A Rapid Data-centric Development Platform for Vision Applications}

\author{
  \hspace{0in}Phoenix X. Huang$^1$
  ~~
  Wenze Hu$^1$
  ~~
  William Brendel$^2$
  ~~
  Manmohan Chandraker$^3$ 
  \\
  \And Li-Jia Li$^4$~~
  Xiaoyu Wang$^1$ 
  \\
  \\
  $^1$ Lighthouse Co.Ltd~~
  $^2$ Heali AI~~
  $^3$ University of California, San Diego~~ 
  $^4$ Stanford University\\
  \hspace{-0in}{\tt\small \{phoenix.huangx, windsor.hwu\}@gmail.com ~ will@heali.ai ~ manu.chandraker@gmail.com}~ \\
  {\tt\small lijiali@cs.stanford.edu ~ fanghuaxue@gmail.com}
}

\begin{document}
\maketitle

\begin{abstract}
This paper introduces an open source platform to support the rapid development of computer vision applications at scale. The platform puts the efficient data development at the center of the machine learning development process, integrates active learning methods, data and model version control, and uses concepts such as projects to enable fast iterations of multiple task specific datasets in parallel. This platform abstracts the development process into core states and operations, and integrates third party tools via open APIs as implementations of the operations. This open design reduces the development cost and adoption cost for ML teams with existing tools. At the same time, the platform supports recording project development histories, through which successful projects can be shared to further boost model production efficiency on similar tasks. The platform is open source and is already used internally to meet the increasing demand for different real world computer vision applications.
\end{abstract}


\section{Introduction}


This paper introduces YMIR\footnote{YMIR denotes You Mine In Recursion. The project is available at: http://www.viesc.com. Code is available at:  https://github.com/industryessentials/ymir.}, a data-centric ML development platform that aims to support developing vision applications at scale. 


YMIR uses data labeling, model training, and active learning based data mining as primitive operations to build datasets and models. It embeds version control into the iterative process to track and manage data annotations and models, much like code version control in software engineering. The platform also introduces the project concept, to isolate different jobs and enable parallel development of datasets for multiple applications.


In terms of system design, YMIR treats dataset and model as the key inputs and outputs of the primitive operations, and stores the specification of each operation into metadata. This allows a decoupled implementation of each operation, which has the following advantages: 1) It allows YMIR to flexibly integrate the best third party tools readily available in the open source community; 2) It reduces the adoption cost for experienced teams, as they can quickly integrate their existing tools into the platform.

YMIR adopts existing version control tools to manage datasets and models, but manages the actual data payload outside the tool to better support frequent export of images to operators as well as frequent updates to the annotations in the dataset development process.


To summarize, the highlights of the YMIR platform are as follows:
\begin{enumerate}
    \item The platform prioritizes data instead of model development in its product design and system architecture.
    \item The platform adopts a modular open design approach to easily integrate third party machine learning tools, and to leverage the contribution of the open source community.
    \item The platform embeds version control of data and model into the system to track the development process.
\end{enumerate}

In the following sections, we will analyze one of the use cases to point out the necessity of prioritizing the dataset development in ML development process, introduce the details of the platform in the aspects of product design and system design, show the key product functions and discuss potential future works.

\section{Background}



Computer vision technology is becoming ready to be used in many industries. For example, object detection alone can help a manufacturing factory to enhance its digital transformation process in many aspects: detect if any worker in the factory is not properly wearing protection gears; check whether all the tools are back into cabinets after a work shift; if the evacuation routes are unintentionally blocked, etc.

Although from many different application verticals, many potential applications share the following traits: the problems can be cast into standard computer vision tasks; the customer has positive image examples, but these are usually rare events buried in their data storage with no annotations; customers do not have a clear cut on the scope of the interested events, so algorithm developing experts need to iterate with end users on task definition through a trial and error process; these applications need to be developed at low cost, as the users are usually hesitate to heavily invest in yet to be verified new technologies.

Due to the challenges described above, a matured tool to semi-automate the active learning based data development process is the key to a viable solution, making project opportunities into real applications. Ultimately, the dataset shall even be curated by the end user directly, as they are the best party to define the task through a labeled dataset. These cases motivate us to develop the YMIR platform, which is useful for developing these applications at scale.



\section{Product Design}\label{produc_design}
\begin{figure}
    \centering
    \includegraphics[width=1\textwidth]{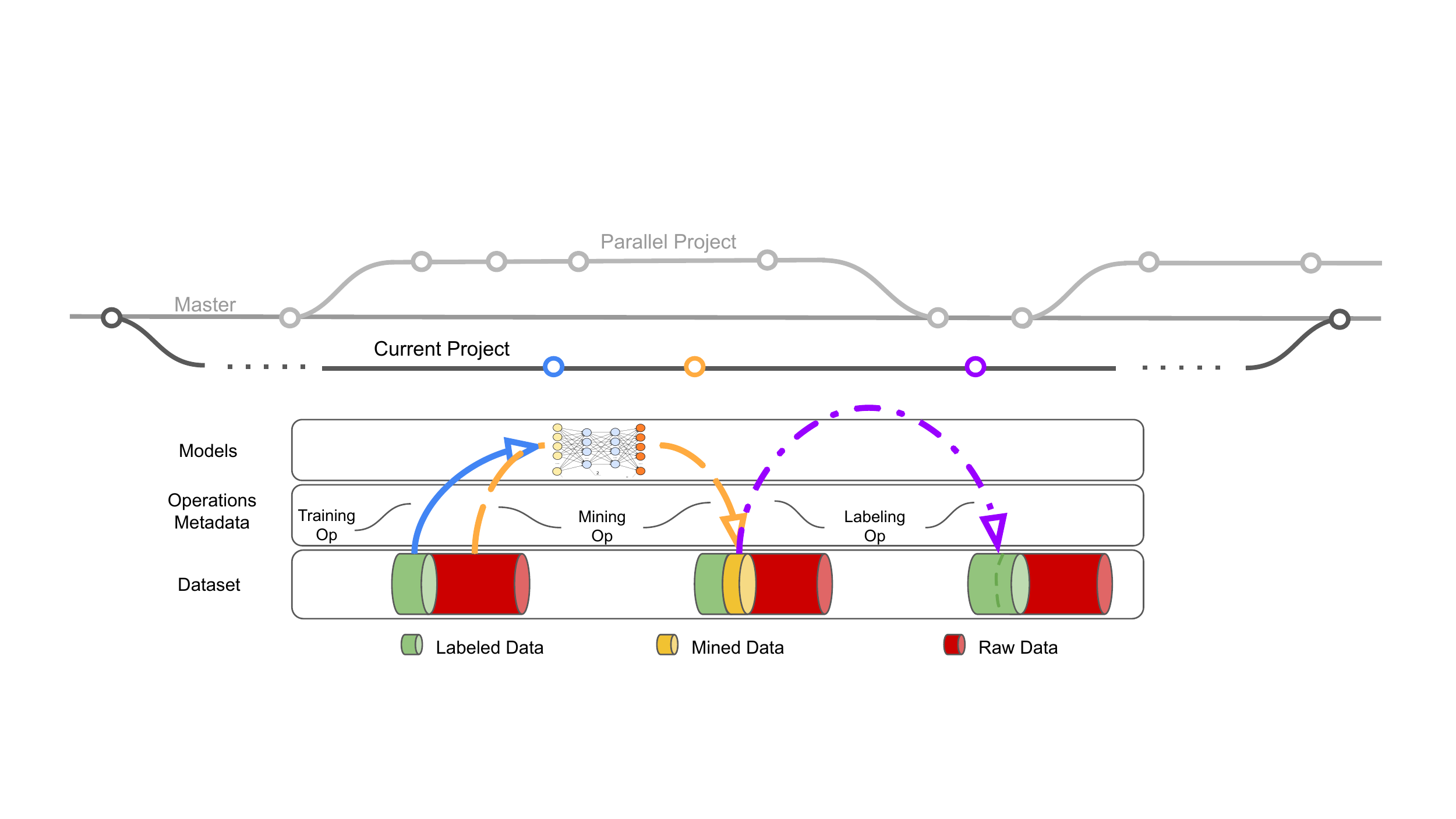}
    \caption{A typical workflow in YMIR. YMIR abstracts the data curation process into states of model, dataset and metadata, as well as the state changing key operations, which are training op, mining op and labeling op. YMIR uses version control to track and mange the data evolution processes.}
    \label{fig:states}
\end{figure}

\subsection{State Space and Operations}\label{data_and_ops}
Following the typical steps in active learning, YMIR models the dataset development process into combinations of 3 primitive operations, which are data mining, model training and data labeling. Correspondingly, the state space of YMIR system is abstracted as the union of datasets, models and metadata. The operations change the states of datasets and models, where the details of the operations are stored in metadata. The relations between these states and operation elements are summarized in Fig.\ref{fig:states}.

\textbf{Dataset.} In YMIR, A dataset is a collection of raw data as assets and the annotations of these assets. We assume most of the assets are image files.  Each dataset has an unique ID as its identifier, and may have a name as a non-unique identifier. Different datasets may share some assets and their annotations.


\textbf{Model.} From the platform's perspective, a model is simply a binary file that can be used to run inferences on input images. Currently, there is no assumptions on the format of the model files, and it is the individual operators which assume the responsibility to check its validity.

\textbf{Metadata.} The metadata stores the types and parameters of the operations applied on the models and datasets which change the states of the system. To accommodate the modularized open design, YMIR only sets very limited number of fields as required in the metatdata, and allows implementations of individual operations specify extra fields if necessary.

\textbf{Mining Op.} The mining op takes a dataset and a model as input, and outputs a subset of the input dataset as a new dataset. The images in the new dataset are usually the most valuable subset to develop a better model. They should be sent to a manual labeling process and later used as training data.

\textbf{Labeling Op.} The labeling op takes an input dataset and outputs an updated version of the dataset with new annotations. YMIR only send images to this op, without any annotations from previous projects. While this design leaves the efforts of merging different annotations to the platform itself, it has practical advantages on the design of data storage and version control, which will be specified in Section~\ref{system}.

\textbf{Training Op.} The training op is expected to take a fully labeled dataset as input and output a new model. For many projects, the scale of the training dataset is usually well below that of COCO dataset, so YMIR trains new models from a fixed starting point such as an imagenet pre-trained model instead of using model checkpoints from previous iterations. This allows users to change model architectures during the development process, which we feel out weighs the benefits of saved computations.

It is worth mentioning that the CRUD (create, read, update and delete) operations on datasets are implemented in the YMIR system instead of via open APIs. This design choice helps hiding the version control concepts out of GUI, thus lowers the barrier for entry level users.

In the training and labeling Op, different datasets may be implicitly merged. The merge takes a primary dataset and one or more secondary datasets, combines their images and image annotations into a new dataset. The new dataset has a new ID, but inherits the name and other auxiliary data from the primary dataset. In terms of conflict resolution, YMIR provides a default implementation that detects annotation conflict using simple rules and overwrite the old versions on conflicting parts. It is still under discussion on whether YMIR should provide interface for custom merging rules as a configurable part in the future versions.



\subsection{Data and Model Version Control}



Proper composition of the defined ops on defined states results in the standard loops of an active learning process, which in theory could result in fast creation of training data and models. YMIR borrows many code version control concepts from git to manage the data and models from this process. Specifically, YMIR uses git branch concept to create new projects so that different tasks on the same set of images can run in parallel. The CRUD of datasets as well as the the primitive operations all create commits to the branch. Each commit stores an updated version of datasets or a new model, as well as the metadata of the operations leading to this change. Finally, only data changes are merged to master branch, which conceptually aggregates all the data annotated through many projects on this platform.

\subsection{GUI and CLI Separation}

As is mentioned in the introduction, YMIR has both a web based GUI and a command line interface (CLI). 

The web interface is designed for efficiently solving typical computer vision problems. Specifically, the GUI only exposes pre-configured primitive operations with default parameters, and hides the concepts of metadata and version control. This is to smooth the learning curve and invite users without in-depth knowledge of machine learning to use our platform for computer vision related development needs.


In contrast, YMIR exposes most of the op parameters together with full functionalities of version control in the command line interface. While a user needs to learn many new concepts, this user interface can unblock expert user's development process for highly customized development requirements.

\section{System Architecture}\label{system}
\begin{figure}
    \centering
    \includegraphics[width=0.6\textwidth]{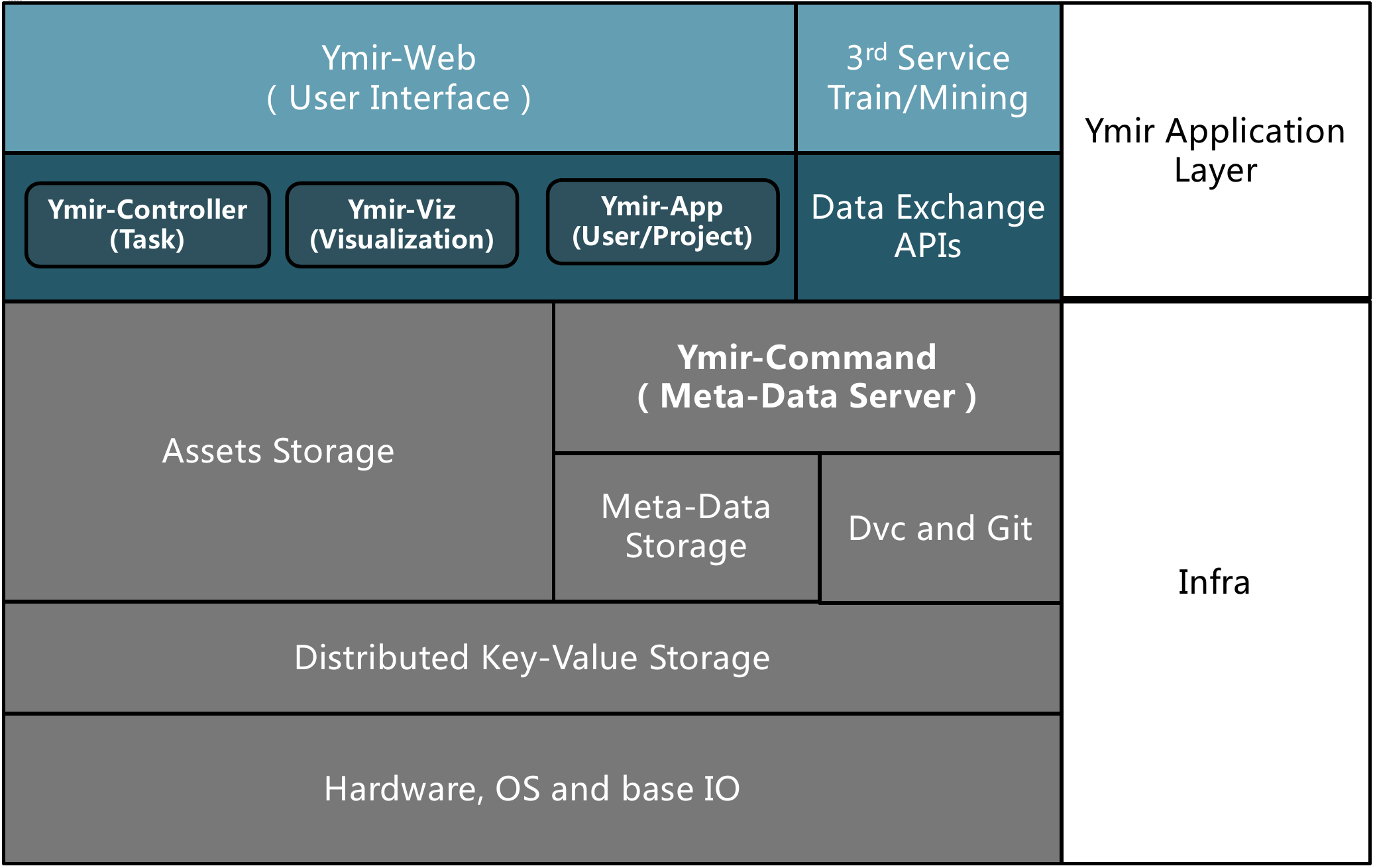}
    \caption{Layered view of the system architecture.}
    \label{fig:system}
\end{figure}


Fig. \ref{fig:system} is an overview of the system architecture. YMIR puts management of system state, \ie datasets and models as the foundation, to support efficient storage and fast retrieval of models, data and versioned data annotations. The application layer is built on top of this foundation, to support flexible expansions of the platform's functions, such as the primitive operations, as well as user and project management. 

\subsection{Dataset and Model Management as Foundations}
YMIR adopted DVC\cite{DVC} for version control of datasets and models, which in turn uses git. On top of DVC, we have made several significant changes to better support fast iterations of datasets:

1. The image assets are moved into dedicated KV storage services, whereas the keys (\ie SHA-1 values of images) instead of images are sent to DVC for version management. We expect data mining operation usually runs on millions images. If the images are stored in DVC, it can easily become bottleneck when exporting these assets to mining or training ops.  

2. The image annotations are stored incrementally. After an image was added to the dataset, its corresponding annotation accumulates along the whole workflow. As the annotation is the most flexible and changeable component, YMIR stores it incrementally as individual packages. The packed annotation is stored in our metadata storage service, and theirs keys are put to DVC for version control. In this case, YMIR reduces the expensive cost of DVC storing the full annotation after each change. This change also helps speeding up exporting the annotations, because usually the training op only use partial annotations related to a target task, which is effectively equivalent to a single change to the annotation. Saving it incrementally allows YMIR to quickly find the related annotation instead of parsing it from the full annotation file.


The metadata storage service is built on top of a standard key value service. Apart from data annotations, this service also stores other metadata: \eg the operation context and parameters (\ie the metadata in Section \ref{data_and_ops}) of each task ops, the attributes of datasets (\ie the dimension, format and keywords of images). YMIR uses protocol buffers to define metadata message format. Fixed metadata (\eg image SHA-hash and attributes) and mutable metadata (\eg task context) are stored separately, which allows to log fixed data in shared packed data part that is built only once, and serialize operation specific data into flexible but low-load storage file that do not affect the enormous fixed metadata.

\subsection{Application Layers}
The modules in application layers can be classified to two categories: the data related primitive operations defined in Section \ref{data_and_ops}, and user related services. 

On the data side, the primitive operations in Section \ref{data_and_ops} are modularized as individual services or processes running in docker containers. As of now, YMIR uses Label Studio \cite{label_studio} as the labeling op, adopts YOLOv4 \cite{yolov4} in DarkNet \cite{darknet} as the training op and the data scoring methods in \cite{active_learning} as the core of our mining op.  

On the user side, the web interface is supported by another layer of services that implements user and project management, op status tracking,  as well as a light weight service to retrieve and visualize the versioned datasets. 


\section{System Functions Overview}

\begin{figure}
    \centering
    \includegraphics[width=1\textwidth]{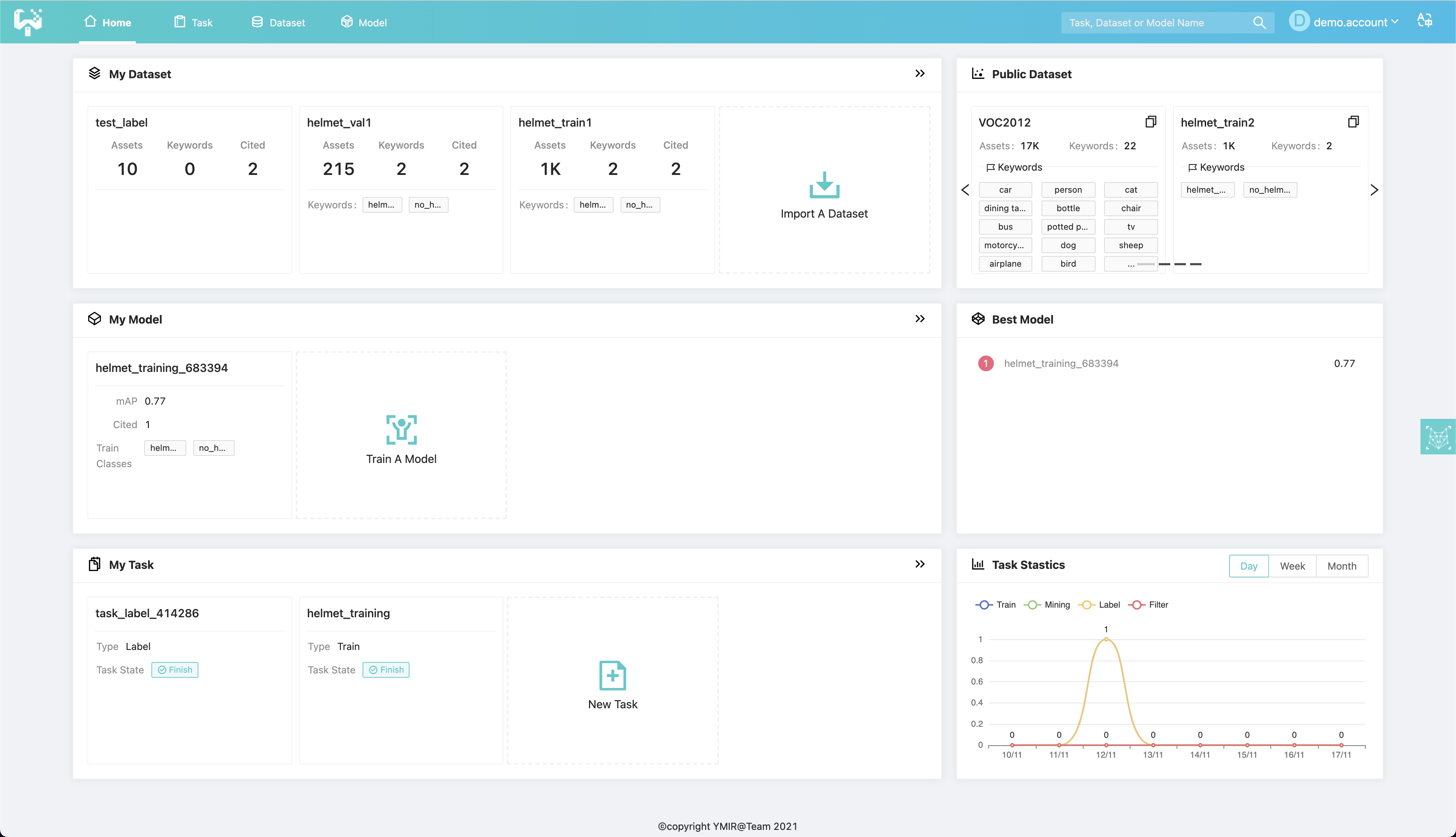}
    \caption{The YMIR homepage.}
    \label{fig::homepage}
\end{figure}

Fig.\ref{fig::homepage} shows the landing page of the YMIR system. Upon login, users have an overview of the datasets, models and ongoing tasks so that users can easily resume working from where it was left off. In the rest of this section, we will briefly list the main functions available in YMIR GUI, as well as the main features inside each function.

\begin{figure}
\centering
\includegraphics[width=0.8\textwidth]{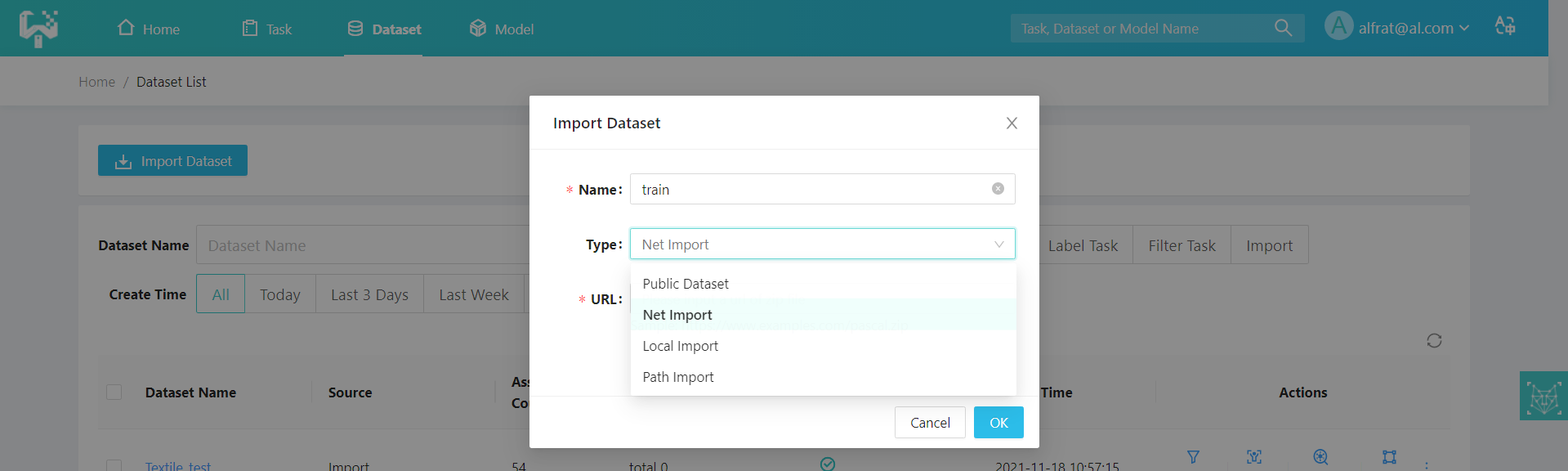}
\caption{UI for dataset import function.}
\label{fig::import}
\end{figure}

\textbf{Dataset Importing.} As is shown in Fig.\ref{fig::import}, YMIR provides three ways to import a dataset: upload dataset, copy from an existing public dataset, and import from an URI. Before importing, the external dataset should be packaged following the PASCAl VOC \cite{voc} dataset format. Also, the administrator of the system can import datasets and set them as public. In this way, a user can easily replicate a dataset by having a shallow copy of it, reducing the time and work needed for packing and importing a dataset. If an user choose to import by URI, the URI can be either a local path, a networked folder address or a http address. 

\begin{figure}
\centering
\includegraphics[width=0.8\textwidth]{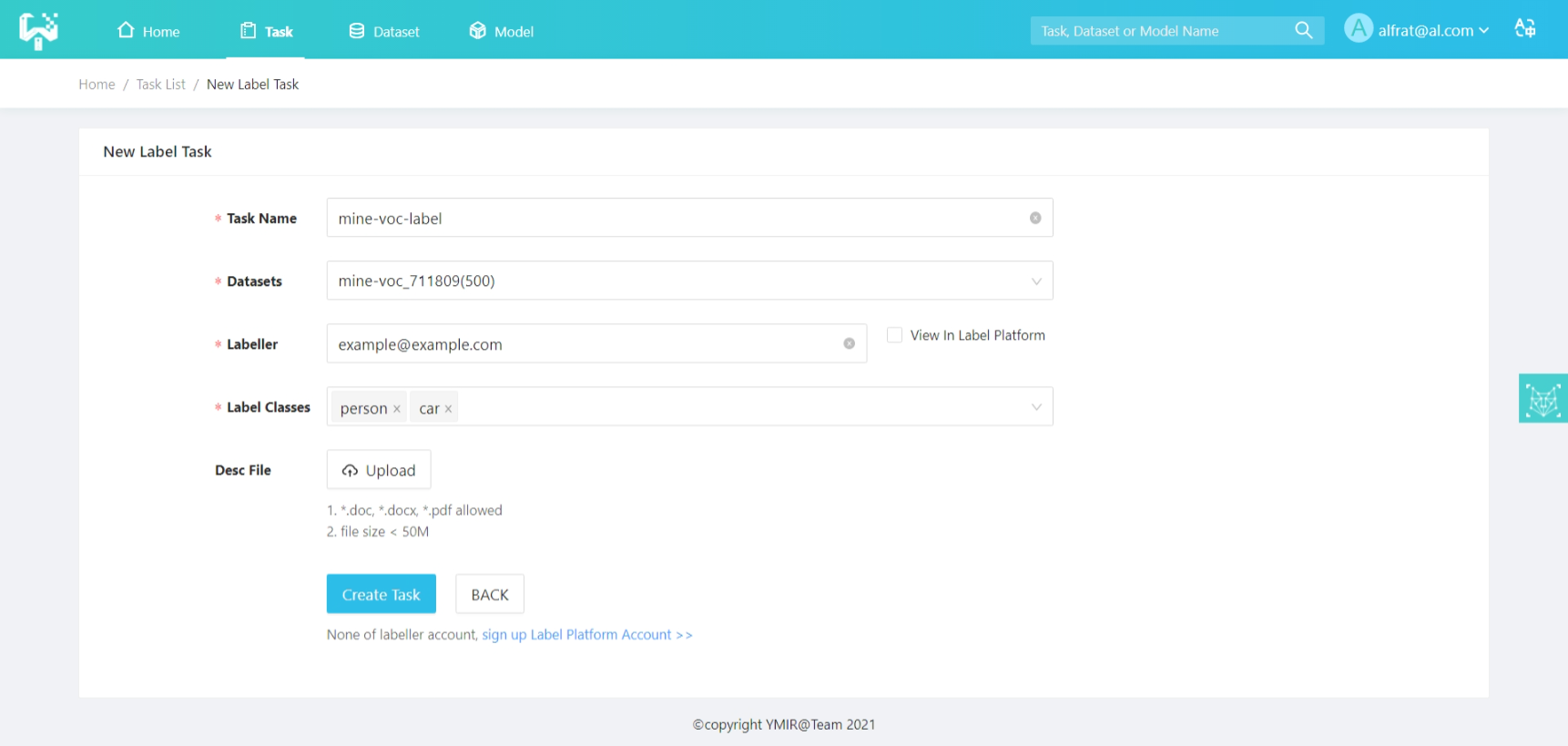}
\caption{Image labeling task creation interface.}
\label{fig::label}
\end{figure}

\textbf{Data Annotations.} After data is imported to the system, users can create annotation tasks through YMIR. YMIR then sends the task to Label Studio, monitors its status, and retrieves labeling results from Label Studio after the annotation task is complete. Although not required, YMIR recommends users to provide a task description file, giving examples of what are the right and wrong labels. A well written document can greatly help raters improve their annotation quality, which significantly improves the accuracy of the resulting trained model. Besides a task description file, users can also specify data annotators through YMIR. This additional feature is based on findings that restricting similar labeling tasks to the same group of raters usually makes the annotation quality much more consistent.

\begin{figure}
\centering
\includegraphics[width=0.8\textwidth]{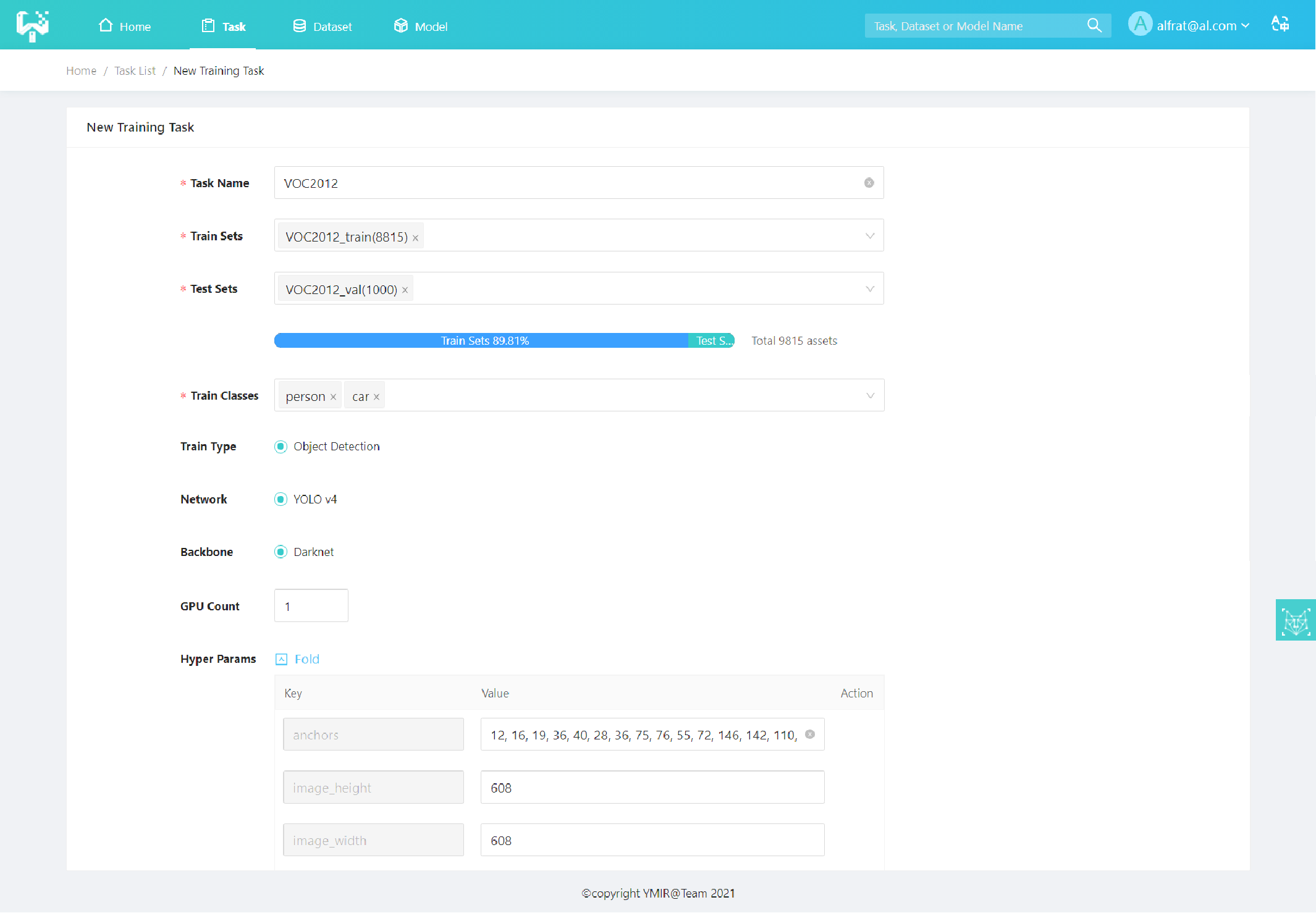}
\caption{UI for model training function.}
\label{fig::train}
\end{figure}

\textbf{Model Training.} Fig.\ref{fig::train} shows the interface for creating a model training task. A task can be created by specifying a task name, the used datasets, as well as the training targets. Note that only object detection task is supported as of the current release. More types of training tasks and more models for each task can be added by both the YMIR developing team and the third party contributors. The UI displays many training related hyper parameters and their default values, which gives the user the flexibility to adjust these details and compare results from different parameter configurations. Once a model is trained, user can see the accuracy metrics on the task result page.

\begin{figure}
\centering
\includegraphics[width=0.8\textwidth]{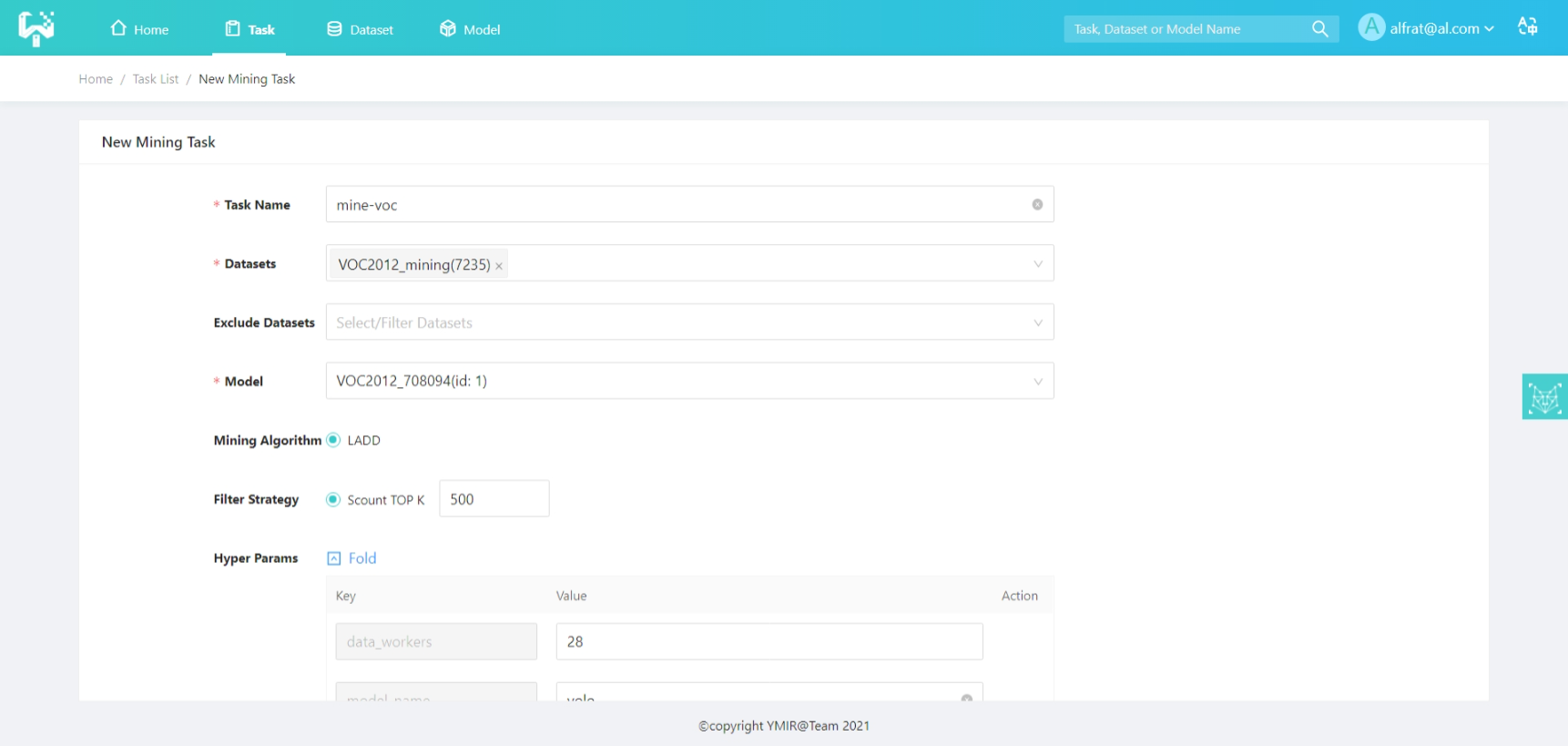}
\caption{UI for data mining.}
\end{figure}

\textbf{Data Mining.} Compared with some existing AI development platforms, data mining is a differentiating feature of YMIR system, and it is the key step in a dataset and model iteration loop. Similar to the model training task, YMIR currently only provides one method for data mining, but exposes many detailed parameters so that interested users can tune the algorithm and compare its results.

\begin{figure}
\centering
\includegraphics[width=0.8\textwidth]{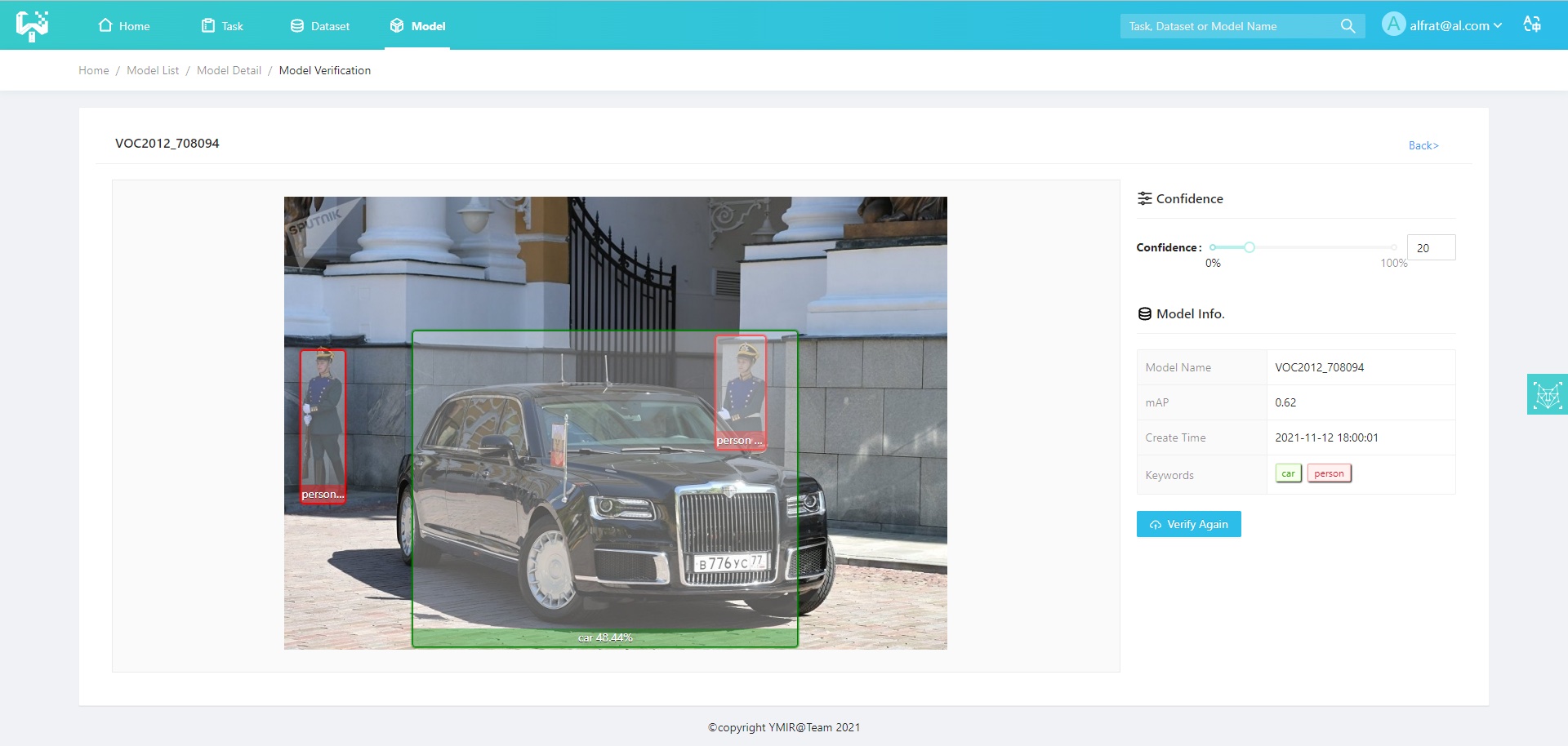}
\caption{UI for model validation.}
\label{fig::validation}
\end{figure}

\textbf{Model Validation.} For any models trained in YMIR system, YMIR provides a model validation feature that lets users test a trained model on the fly. Through the interface, users can upload an image and check the detection results on that image in seconds. Fig.\ref{fig::validation} shows an example detection result of an image on a trained detection model. 

\begin{figure}
    \centering
    \includegraphics[width=0.9\textwidth]{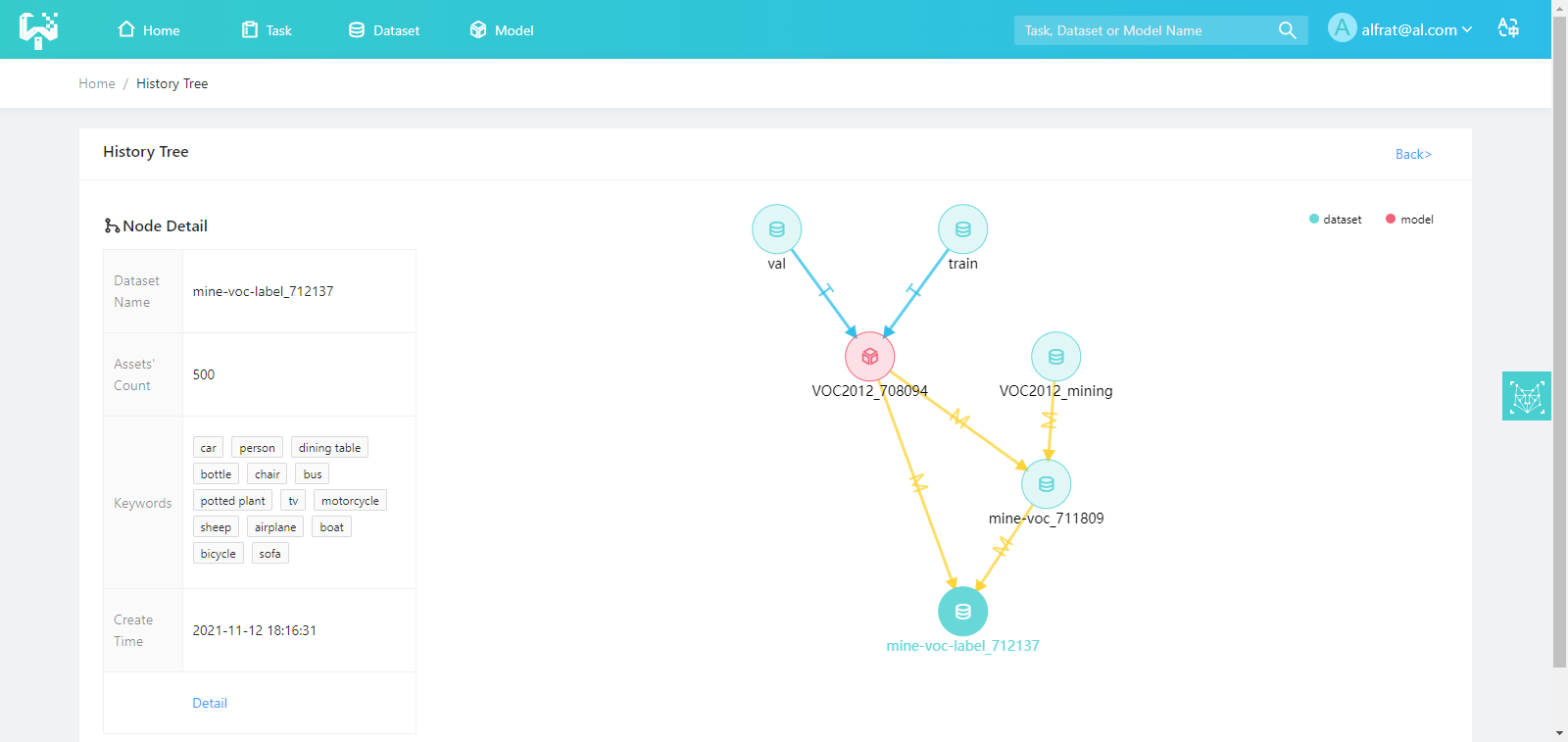}
    \caption{Model/Dataset history tracking.}
    \label{fig::history}
\end{figure}

\textbf{Model and Dataset History} The model and dataset history page (Fig.\ref{fig::history}) shows the dependency relations between datasets and models, as models can be used to mine new datasets, and new datsets can then be used to train new models. It provides a story-line view of the entire model and dataset development process, which may provide clues on discovering the good and bad practices in the ML development process.

\section{Summary and Disucssions}
This paper presents a new platform to support efficient development of computer vision applications at scale. Before made open source, the platform was adopted by pilot users, who reported significant productivity boost compared to using standard manual process.

Although not fully utilized yet, the development histories stored in this platform may further serve as training data to optimize the productivity of users. If widely adopted, sharing and exchange of these records may further help the community to find even better ways to solve data centric AI problems.


{\small
{
\small
\bibliographystyle{plain}

}

}

\end{document}